# A New Approach for Super resolution by Using Web Images and FFT Based Image Registration

Archana Vijayan[1], Vincy Salam[2]

[1]P.G Scholar, Department of Computer Science, Musaliar College of Engineering and Technology, Pathanamthitta, India
[2] Professor, Department of Computer Science, Musaliar College of Engineering and Technology, Pathanamthitta, India

*Abstract*- **Preserving accuracy is a challenging issue in super resolution images. In this paper, we propose a new FFT based image registration algorithm and a sparse based super resolution algorithm to improve the accuracy of super resolution image. Given a low resolution image, our approach initially extracts the local descriptors from the input and then the local descriptors from the whole correlated images using the SIFT algorithm. Once this is completed, it will compare the local descriptors on the basis of a threshold value. The retrieved images could be having different focal length, illumination, inclination and size. To overcome the above differences of the retrieved images, we propose a new FFT based image registration algorithm. After the registration stage, we apply a sparse based super resolution on the images for recreating images with better resolution compared to the input. Based on the PSSNR calculation and SSIM comparison, we can see that the new methodology creates a better image than the traditional methods.**

*Keywords:* **SIFT extraction, FFT, Super resolution, Image retrieval, Bicubic Interpolation**

## 1. INTRODUCTION

Image super-resolution (SR) is not an easy task as one low resolution (LR) image might relate to different high resolution (HR) images. Hence, we need to impose some constraints on the original images to make sure that the image SR is well tractable. Primarily, these rules make use of the correlations within the image or across different images. According to this method, image SR can be categorized into three: exemplar-based, multi-image-based, and interpolation-based. The interpolation-based methods [2] which do not use further related information might face problems from blurring artifacts since it is difficult to infer HR details when provided with only one LR Image. Multi-image-based methods combine different LR images of the same view to give further details for retrieving HR details [3]. Methods which use multiple images normally give better results compared to the ones which uses single image. However, it is still not to retrieve high-frequency (HF) information in case of high values of magnification factor [3]. Exemplar-based methodology [4] prepares a training set of HR/LR pair sections to infer HR sections from LR sections of the images. With the help of external HR images for extra information, this method can generate new and believable HR information. However, SR performance is limited by the fact that the training set is normally fixed.

Various studies suggest that the most important ingredient for achieving high SR performance is to make use of the correlation from other available images to retrieve the HR details in an effective manner. Due to the advance in image processing techniques, highly potential image retrieval methods are available to extract correlated pictures from the World Wide Web. These methods work by correlating matching regions in the retrieved pictures. However, the fact that the different retrieved pictures of the same view would have been captured from different angles by using different devices of varying resolutions makes it difficult to find matching sections without aligning them properly. These limitations reduce SR performance improvement while we use images retrieved from internet as assisted information. Amongst the vast collection of images, we can still find a group of images which are highly correlated, which are taken at landmark places across the world that can be called as landmark images. The Internet boom and the increasing popularity of social networking sites has increased the availability of such landmark images to a great extent. For such an image, we can easily find a lot of correlated images which are taken with the help of cameras with varying resolution from different angles. If we can ignore the different in resolution and view points, we can call these images as partial duplicates of each other. With the help of a geometric transformation, we can align the images with the query image easily. This will facilitate considerably high performance SR from the retrieved images.

As per the above observations, we are proposing a method of hallucinating landmark images using highly correlated pictures from the World Wide Web. However, implementation of this method is not an easy task. First of all, Searching of HR images with the help of the input LR image is not easy. Existing image retrieval methods are designed for images with matching resolution whereas the images available on the internet will not give us this luxury. Also, hallucinating realistic objects is not easy even if we have highly correlated images since they might be taken from different angles and they might contain objects which are irrelevant. We should also keep in mind that we still have to generate a better and reliable image even in the absence of correlated images.

Aim of this work is to solve the above problems using a method comprising of 3 steps. In the first step, a highly correlated high resolution image dataset is created based on the features of the input image. As discussed above, these images in the dataset will be of different viewpoints, resolution and illumination. In the second step, with the help of FFT based global registration and a structure-aware criterion of matching patches, the accuracy of matching between the LR and HR images is improved. A patch based image SR is used in the third step to build the HR image.

## 2. EXISTING SYSTEM

In the existing system [1] a High resolution image is created from landmark images by implementing global registration and SIFT. A bicubic interpolation algorithm is used to interpolate the input image. With the help of Scale invariant feature transform method, SIFT feature points are identified on the image. Identification of SIFT feature point of all the images in the test database using the same method happens in parallel. A threshold value is set for the number of matching feature points and all the images which are having matches beyond this threshold are identified. These images and the input image could differ in terms of its color, focal length, texture ,orientation etc. All these images will be aligned to the input image using global registration in the next step. Correlation of the above images is used to create a high resolution image during the final phase. An explanation of these steps is given in this section.

The existing system recovers HF details from a training set consisting of HR/LR pairs. The up-sampled image of the input image is split into overlapping sections and each of these sections is searched against the training set to retrieve matching parts. The high resolution section from the retrieved image patch will be then added to the input image to create the HR image. But the matched patch should be similar to the input patch and the retrieved HR patch should create consistency while adding to the original input image. C. Hsu and C. Lin propose a method of adopting scale-invariant feature transform (SIFT) descriptors to improve the training set [10]. In order to comprehend reasonable details in image regions, high-level image analysis is integrated [11] with low level image synthesis. The input LR image is considered as small textures and each of these textures is matched against relevant textures within the database. A context-constrained recreation method is discussed in [11] to enhance the SR performance. The results of this method demonstrate the potency of finding matched sections in a similar object set.

Image retrieval is a familiar application on the World Wide Web. Partial or close to duplicate image retrieval has drawn wide attention as it is useful for many applications. Local feature descriptors are used widely in partial or close to duplicate extraction. For each SIFT descriptor, quantization is done to the nearest visual patch and then used for similarity rating. This greatly speeds up image retrieval but at the same time decreases the power of SIFT descriptors. To increase the accuracy, each SIFT feature is quantized to a weighted set of visual patches instead of a single patch to reduce the effect of the quantization error. Performance of this method is enhanced by applying Hamming embedding to the descriptors and integrating weak geometric consistency inside the inverted file system.

This system aligns the matching images with the help of SIFT-based global registration. The output images are highly matching but with considerable difference at the pixel-level as they are created with different viewpoints, illumination and focal lengths. The HF correlation will be less if we implement direct SR from these images. Since it is generated through pixel-wise matching, even when very similar images are provided, the HF information inferred at the patch level is insufficient. An approximate alignment is achieved via geometric global registration and this is further polished by local patch-based matching. Since the SIFT descriptors of all the matching images are available, the registration transformation among the ULR image and one matching image is obtained by exploiting the Random Sample Consensus (RANSAC) algorithm**.**

## 3. PROPOSED SYSTEM

### A. *Problem Definition*

The existing system uses a method called Global Registration to align the different matching images which are retrieved from the image database. This method has substantial limitation when it comes to complex images such as face and animals. Also, for images containing rich textural regions without dominant structures, such as grass and leaves, which are hard to register using the SIFT descriptors; this method of registration is not accurate. Also, in the existing system, exemplar based image SR is used to build high resolution images from the registered images. Though, this method creates a better image in terms of resolution compared to the input image, the noise percentage is still high. Also, this method may not always provide a better structure similarity between the input and the output images. So, our aim is to find a better method to overcome the above limitations. This could be achieved by employing FFT based image registration and a better SR algorithm to produce high resolution value with the weighted average of the two translated values on both sides.

B. Solution Description

There are seven modules in the proposed system. The modules are as follows:
1. Create database
2. Bicubic Interpolation
3. Feature Extraction
4. Matching
5. Image Registration
6. Super Resolution
7. Comparative Analysis

*1) Create database*

Create database is the first module used in the proposed system. We need to make a good collection of images for this. Images can be collected from various sources from internet like Google, Flicker etc. We are planning to use images based on the Table 1 given below

Table 1 Image collection

| Country | Places |
|---------|--------|
| China | Summer Palace, Great Wall, Forbidden city |
| Egypt | Aswan Temple, Luxor Temple, Sphynx |
| Japan | Kinkakuchi Temple, Kiyomizu Temple, Nijo Castle, Ryoanji Temple |

*2) Bicubic Interpolation*

Interpolation is the process of transforming image from one resolution to another resolution without compromising image quality. Image interpolation is very important function in the field of Image processing, for doing enhancement of image, zooming, resizing and many more. In arithmetic, bicubic interpolation is an enhancement of cubic interpolation for interpolating data points on a regular grid with two dimensions. The resulting surface is smoother than corresponding surfaces obtained by traditional methods like bilinear interpolation or nearest-neighbor interpolation. Bicubic interpolation can be achieved using cubic splines, Lagrange polynomials, or cubic convolution algorithm. In image processing, bicubic interpolation is normally chosen over bilinear interpolation or nearest neighbor interpolation for image re-sampling, when speed is not an important constraint. Unlike bilinear interpolation, which only takes 4 pixels (2×2) under consideration, bicubic interpolation considers 16 pixels (4×4). Images re-sampled with bicubic interpolation are normally smoother and have less number of interpolation artifacts. In bicubic interpolation, the block uses the weighted average of four converted pixel values for each output pixel value to create the output matrix by replacing each input pixel

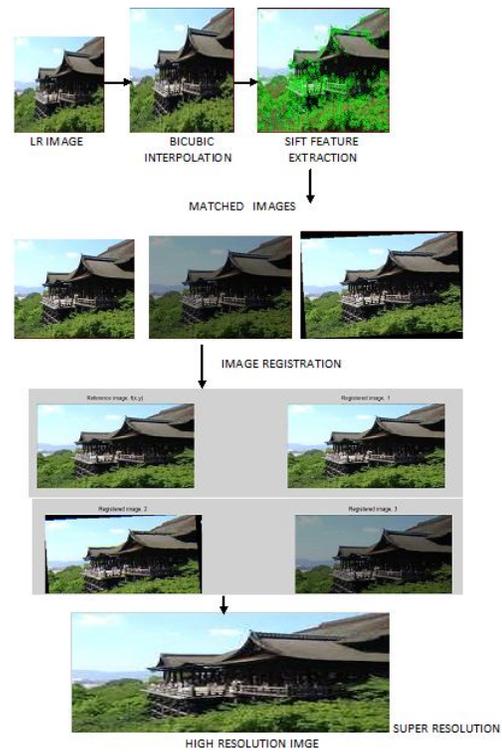

Fig 1 Framework of Proposed System

When using the bi-cubic interpolation zoomed image intensity(x,y) is defined using the weighted sum of mapped 16 neighbouring pixels of the orginal image which is same as in the above method. Let the zooming factor is 's', and the mapped pixel point in the orginal image is given by 'r' and 'c'. Then the neighbor-hood matrix can be defined as

$$\begin{pmatrix} p_{11} & p_{12} & p_{13} & p_{14} \\ p_{21} & p_{22} & p_{23} & p_{24} \\ p_{31} & p_{32} & p_{33} & p_{34} \\ p_{41} & p_{42} & p_{43} & p_{44} \end{pmatrix} =$$
$$\begin{pmatrix} (f(r-1,c-1)) & (f(r-1,c)) & (f(r-1,c+1)) & (f(r-1,c+2)) \\ (f(r,c-1)) & (f(r,c)) & (f(r,c+1)) & (f(r,c+1)) \\ (f(r+1,c-1)) & (f(r+1,c)) & (f(r+1,c+1)) & (f(r-1,c+2)) \\ (f(r+2,c-1)) & (f(r+2,c)) & (f(r+2,c+1)) & (f(r+2,c+2)) \end{pmatrix}$$

.(1)

Using the bi-cubic algorithm;

$v(x,y) = \sum_{i=0}^{3} \sum_{j=0}^{3} a_{ij} \; p_{ij}$ ……………………… (2)

The coefficients $a_{ij}$ can be find using the La-grange equation.

$a_{ij} = a_i * b_j$ ………………………………… ………. (3)

$a_{i} = \prod_{k=0, k \neq l}^{3} \frac{(x - cell(s*(c+k)))}{cell(s*(c+l)) \; _{-} \; cell \; (s*(c+k))}$ ……… … (4)

$$b_{j=\prod_{k=0, k\neq 1}^{3}} \frac{(y-cell(s*(r+k)))}{cell(s*(r+j))\_cell(s*(r+k))} \quad \ldots\ldots\ldots \quad (5)$$

When implementing this algorithm a mask for defining ai and bj is made using matrix and then applied it to the matrix which is having the selected 16 points, in order to reduce the complexity of the algorithm and to reduce the calculation time. Zero padding is applied to envelop the original image to remove the zero reference error occurred.

$$v(x,y) = \{a_1 \quad a_2 \quad a_3 \quad a_4\} * \begin{pmatrix} p_{11} & p_{12} & p_{13} & p_{14} \\ p_{21} & p_{22} & p_{23} & p_{24} \\ p_{31} & p_{32} & p_{33} & p_{34} \\ p_{41} & p_{42} & p_{43} & p_{44} \end{pmatrix} * \begin{bmatrix} b_1 \\ b_2 \\ b_3 \\ b_4 \end{bmatrix}$$

$$\ldots\ldots (6)$$

*3) Feature extraction*

Scale-invariant feature transform (or SIFT) is an algorithm used to identify and depict local features in images. The SIFT algorithm takes an image as input and transforms the image into a compilation of local feature vectors. Each of these vector attributes is believed to be distinguishing and invariant to any scaling, translation or rotation of the image. In the implementation, these features are used to find distinguished objects in different images and the transform can be extended to match objects in images.

This algorithm is one of the widely used algorithm for image feature extraction. SIFT retrieves image features that are stable over image transformation, rotation and scaling and considerably invariant to changes in the illumination and camera angle. The SIFT algorithm has four major phases. They are Key point Localization, Extrema Detection, Orientation Assignment and Key point Descriptor Generation as given in Fig 2. Extrema Detection examines the image under different scales and octaves to segregate points of the image that are different from their background. These points, called extrema, are probable candidates for image features. The next phase, Key point Detection, begins with the extrema and identifies some of these points to be key points, which are a whittled down a group of feature candidates. This enhancement rejects extrema, which are induced by edges of the picture and by points of low contrast. The third phase, Orientation Assignment, transforms each key point and its surrounding points into a set of vectors by calculating a magnitude and a direction for them. It also finds out other key points which could have been missed in the first two phases; this is done on the basis of a point having a considerable magnitude without being an extremum. The algorithm has now identified a final set of key points. The final phase, Key point Descriptor Generation, takes a group of vectors in the neighborhood of each key point and consolidates this data into a group of eight vectors called the descriptor. Each descriptor is transformed into a feature by calculating a normalized sum of these vectors.

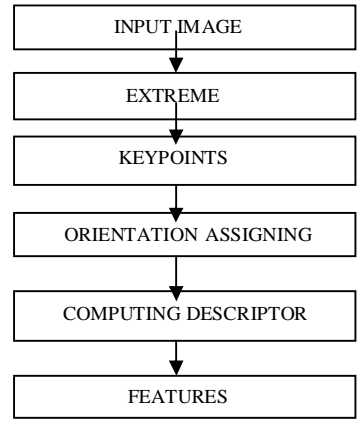

Fig 2 Major phases of SIFT algorithm

SIFT descriptors differentiate image regions invariant to image rotation, scale and 3D camera viewpoint. They are highly distinguished and thus extensively used in image retrieval, 3D restoration, and panoramic stitching. A SIFT descriptor is denoted as

$$f = \{v, x, s, o\} \quad (1)$$

where v is a 128-dimension vector denoting one image region by gradient histograms in different directions. The scale, location, and dominant orientation are represented by x, s, and o in that order. A SIFT descriptor is retrieved from a group of scale space images created from an input image by both Gaussian filtering and down-sampling hierarchically. The feature scale and location are identified by the maximum and minimum of difference-of-Gaussian space, which is the difference between two neighboring Gaussian blurred images. One or more orientations are computed in a region that is centered at the location of the feature and the size is determined by the corresponding scale. The vector covers the gradient information computed from 16 sub-regions after the region is altered according to the prevailing orientation. We note that an image is down-sampled into various octaves during SIFT generation. Though SIFT descriptors are invariant to image scale, SIFT descriptors retrieved from the input LR image are different from those retrieved from the original HR image. In our method, we estimate the HR details from the correlated HR images extracted by the LR image. The correlated images are identified by matching the SIFT descriptors among the up-sampled LR (ULR) image and the HR images from web.

*4) Matching*

Usage of a fixed training set restricts the ability of exemplar-based SR in finding highly correlated HR patches. Our method uses an adaptively retrieved HR image collection from a large-scale database created by crawling images from the Internet. Bundled SIFT descriptors are used for image retrieval in our method, which are more distinguished than a single descriptor in a large scale partial-duplicate web image search. In the bundled recovery, the SIFT descriptors are classified according to spatial correlations between them. One image region coupled with a large-scale descriptor normally contains a few small-scale descriptors. These descriptors are bundled as one set to serve as a primary unit in feature matching. To step up the search process, the BoW (Bag-of-Words) is used in which the SIFT descriptors are accumulated into visual words. The bundled sets of one image are then saved as one inverted-file index, as demonstrated in Fig. 4. Each visual word has an entry in the index which consists of the list of images in which the visual word is presnt. It also has the number of members in the set centered on this visual word, which is followed by the members of visual words and sector index. The number of member visual words in a set is restricted to 128 and the count of indices is set at 4. The SIFT descriptors retrieved from the query image ~ I are also grouped into groups in the same way. Then each of these bundled set is matched against the bundled sets saved in the inverted-file index. The matching is scored by the amount of matching visual words and the geometric relationship. Then this score is mapped to the image associated with the matched group. Once all the bundled sets in the query image are matched, the correlation among a candidate image and the query image is calculated by the sum of the scores of matched bundle groups among them. A higher total score suggests a better correlation. Images with many highest total scores are preferred as correlated images for the query image. The visual group based retrieval normally performs better than the Bow method in terms of the mean average precision (mAP) owing to the usage of spatial correlations between descriptors. The below figure represent the retrieval query.

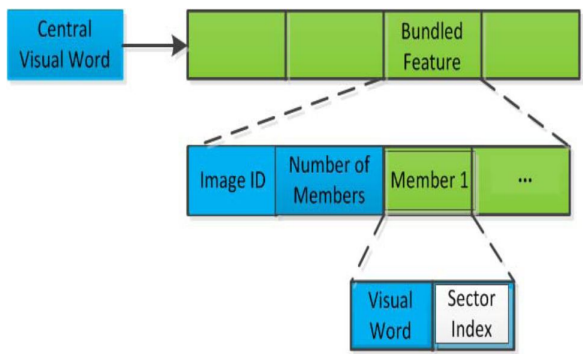

Fig 3   The inverted file index used in our retrieval scheme

*5) Image registration*

Image registration is the process of converting different group of data into one coordinate system. Data may be different photographs, data from various times, depths, sensors or viewpoints. It is applied in computer vision, biological imaging, medical imaging, brain mapping, military automatic target recognition, and assembling and analyzing images and information from space satellites. Registration is essential in order to be able to evaluate or integrate the information recived from these different sources. Frequency-domain approaches find the conversion parameters for registration of images while participating in the transform domain. Such approaches work for simple conversions, such as rotation, translation, and scaling. Applying the phase correlation methodology to a pair of images creates a third image that contains a single peak. The position of this peak corresponds to the relative translation between the various images. Unlike most spatial-domain algorithms, phase correlation methodology is resilient to occlusions, noise, and other defects which are common in medical or satellite images. Furthermore, the phase correlation makes use of the fast Fourier transform to calculate the cross-correlation among the two images, usually resulting in large performance advantages. The methodology can be extended to measure rotation and scaling differences among two images by first transforming the images to log-polar coordinates. Owing to the properties of the Fourier transform, the scaling and rotation parameters can be measured in a manner invariant to transformation.

*Algorithm for FFT based image registration*

1. Provide one reference image and three matched images
2. Use image registration via correlation
3. Compute error for no pixel shift
4. Compute cross correlation by an IFFT and locate    the peak
5. Up-sample by a factor of 2 to obtain initial estimate
6. Embed Fourier data in a 2x larger array
7. Compute cross correlation and locate the peak
8. Obtain shift in original pixel grid from the position of the cross correlation peak
9. If up sampling > 2, then refine estimate with matrix multiply DFT
10. Locate maximum and map back to original pixel grid
11. Compute registered version of first image.
12. Continue the above steps for the other two images

*6) Super Resolution*

Here we are introducing a new method to single-image super resolution, primarily based upon sparse signal representation. Analysis on image statistics suggests that we can represent image patches as a sparse linear amalgamation of elements from an aptly chosen over-complete dictionary. Encouraged by this observation, we look for a sparse representation for each patch of the low-resolution input image, and then use the coefficients of this representation to create the high-resolution output image. Hypothetical results from compressed sensing recommend that under mild conditions, the distributed representation can be correctly retrieved from the down sampled signals. By conjointly training two dictionaries for the low- and high-resolution image sections, we can implement the similarity of distributed representations between the low-resolution and high-resolution image section pair corresponding to their own dictionaries. Hence, the distributed representation of a low-resolution image section can be applied with the high-resolution image patch dictionary to create a high-resolution image patch. The learned dictionary pair is a more perfect representation of the patch pairs, in comparison with the previous approaches, which plainly sample a large number of image patch pairs, reducing the computational cost to a great extent. The usefulness of such sparsity prior is demonstrated for both universal image super-resolution (SR) and the particular case of face hallucination. In both these cases, our algorithm creates high-resolution images that are viable or even better in quality to images formed by further similar SR methods. Furthermore, the local sparse modeling of our method is generally robust to noise, and hence the proposed algorithm can handle SR with noisy inputs in a more combined framework.

*Algorithm for super resolution*

1. Input: training dictionaries Dh and Dl, a low resolution image y.
2. Normalize the dictionary
3. Solve the optimization
4. Bicubic interpolation of the low resolution image.
5. Extract low resolution image features.
6. Recover low resolution patch
7. Generate the high resolution patch and scale the contrast.
8. Fill in the empty with bicubic interpolation

### 4. CONCLUSIONS

We propose a novel method to hallucinate landmark images by retrieving correlated web images. In contrast to the exemplar-based hallucination method in the existing system, we use a patch based super resolution method to build an adaptive correlated image set for each LR image by retrieving correlated images from the Internet. In the global registration stage, these images are registered with the LR image. Therefore, we can retrieve more accurate high frequency details from the registered images.
.

### 5. EXPERIMENTS

Here we evaluate the performance of our proposed SR scheme. We first present the details of the image database generated from the Internet. Then the SR results are evaluated both objectively and subjectively in comparison. Finally, we discuss the impact of illumination and correlated images, and analyze the computational complexity of our algorithm.

Table 2 comparative analysis

| IMAGE | BASIC | | PROPOSED | |
|---|---|---|---|---|
| | PSNR | SSIM | PSNR | SSIM |
| A | 26.53 | 0.9008 | 27.26 | 0.91651 |
| B | 27.58 | 0.8805 | 27.61 | 0.899 |
| C | 26.42 | 0.8201 | 27.01 | 0.8473 |
| D | 25.32 | 0.9232 | 26.01 | 0.9721 |
| E | 27.53 | 0.8332 | 27.81 | 0.8523 |

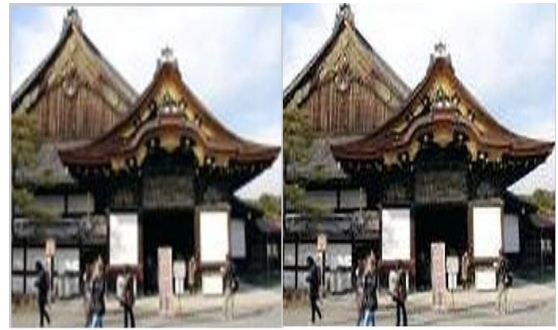

Fig 4    (a)basic            (b) proposed

Table 2 shows the comparison between our method and a existing method. The results clearly show that our method is more efficient. Figure 4 shows similarity and difference between the output images produced by the existing and the proposed systems Output image produced by our methodology has high PSNR value which proves that the output has less noise compared to the existing system. A comparison of SSIM is also demonstrated in table 2 which supports our claim that our method produces better result than the existing system.